\PassOptionsToPackage{table}{xcolor}
\PassOptionsToPackage{pagebackref,breaklinks}{hyperref}

\documentclass[]{bytedance_seed}

\usepackage[toc,page,header]{appendix}

\input{preamble}

\newcommand{\ours}{MetaPoint\xspace}

\usepackage{enumitem}
\newtcolorbox{mybox}[1]{%
  colback=gray!10,
  colframe=gray!50,
  coltitle=black,
  fonttitle=\bfseries,
  boxrule=0.5pt,
  arc=2pt,
  title=#1,
  width=\textwidth,
}
\usepackage{listings}
\lstdefinelanguage{json}{
    basicstyle=\ttfamily\small,
    breaklines=true,
    showstringspaces=false
}

\usepackage{colortbl}
\usepackage{amsmath}
\usepackage{bbding}

\title{MetaPoint: Unlocking Precise Spatial Control in Agentic Visual Generation}

\author[1,2,*]{Dewei Zhou}
\author[2,*]{Xinyu Huang}
\author[2,*]{Xun Wang}
\author[1,2]{Ji Xie}
\author[2]{Yabo Zhang}
\author[2]{Liang Li}
\author[2]{Kunchang Li}
\author[3]{Zongxin Yang}
\author[1,\dagger]{Yi Yang}

\affiliation[1]{Zhejiang University}
\affiliation[2]{ByteDance Seed}
\affiliation[3]{Harvard University}

\contribution[*]{Equal contribution}
\contribution[\dagger]{Corresponding author}
\contribution[]{Xun Wang is the project leader}

\abstract{
Generative visual models fundamentally struggle with precise spatial control. This arises from a core disconnect: models can process textual descriptions of space but cannot directly map numerical coordinates onto the 2D image canvas (as illustrated in Fig.~\ref{fig:sota_cases}). We introduce \textbf{\ours}, a method that bridges this gap by representing a continuous 2D coordinate as a single, special token. Crucially, \ours requires no new architectural components; it directly leverages the model's inherent positional encoding schemes to interpret these coordinates, treating our token as a virtual point on the canvas. This lightweight approach enables pixel-level control of an object's position with one token or its bounding box with two, all without requiring architectural changes or bespoke attention masking. The \ours tokens are designed to be compositional, serving as spatial primitives. This allows a planner agent to decompose a high-level user request into a structured sequence of primitives for the generator. By providing a simple, precise, and scalable building block for spatial control, \ours unlocks more powerful compositional generative agents and enables intuitive, interactive editing systems.

}

\date{\today}
\correspondence{Yi Yang at \email{yangyics@zju.edu.cn}}

\begin{document}
\maketitle

\begin{figure*}[t]
\centering
\includegraphics[width=0.9\linewidth]{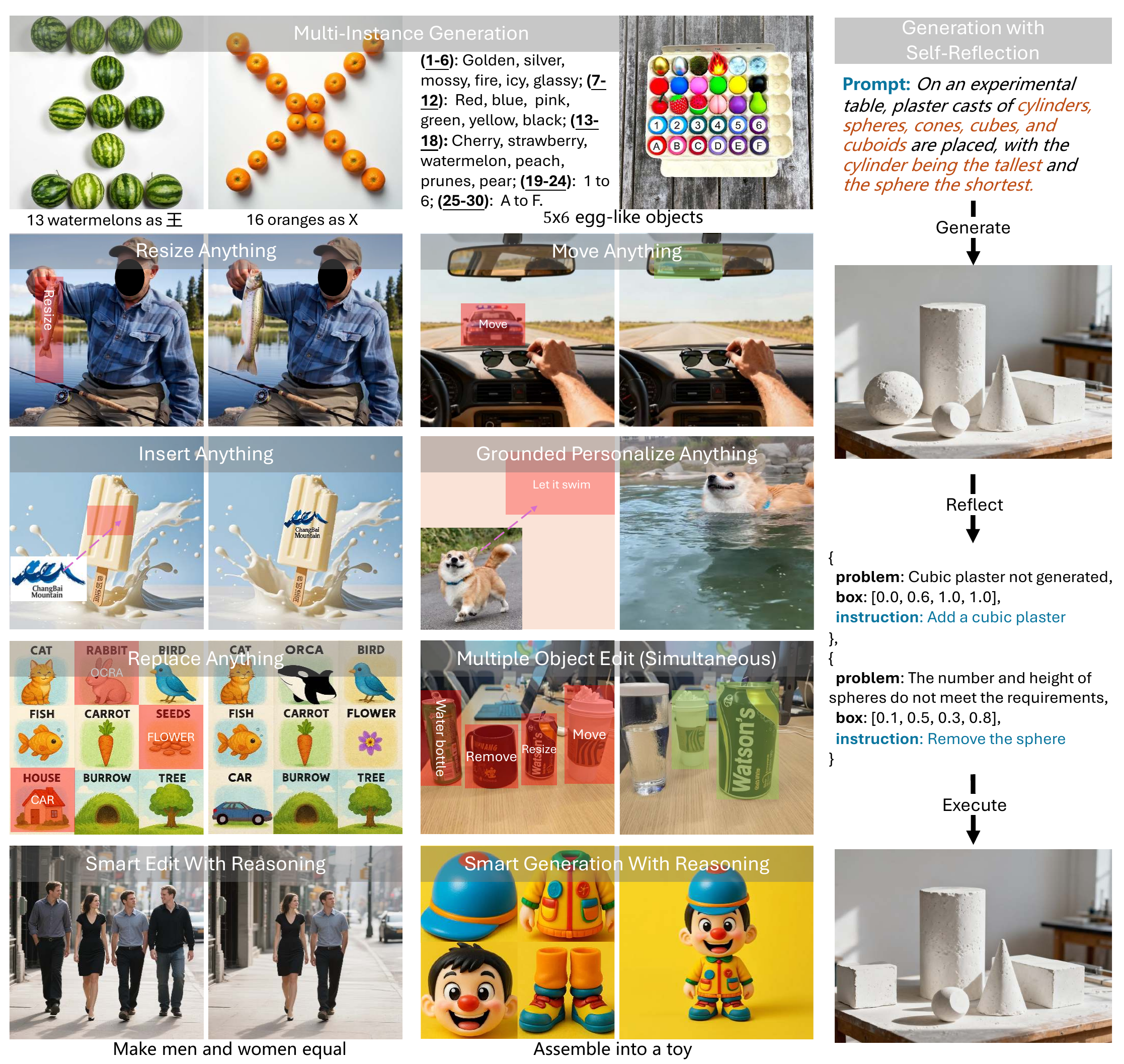}
\vspace{-2mm}
\caption{
\textbf{\ours} demonstrates a wide array of capabilities, including complex \textbf{multi-instance generation}, versatile object editing (\textbf{move}, \textbf{insert}, \textbf{resize}, \textbf{replace}, and \textbf{multi-object editing}), and high-level \textbf{smart generation} with reflection from VLM-agent.
}
\label{fig:teaser}
\end{figure*}

\section{Introduction}
\label{sec:intro}

\begin{figure}[!t]
   \includegraphics[width=\linewidth]{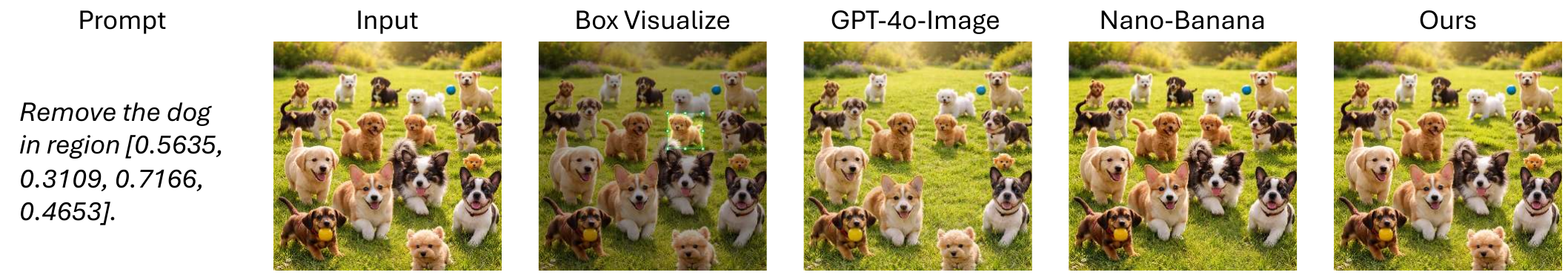}
   \vspace{-2mm}
\caption{Given a coordinate-based editing instruction, GPT-4o-Image~\citep{gpt-4o} and Nano-Banana~\citep{nanobanana} both fail to precisely follow simple coordinates, exposing a key barrier to exact grounded generation. In contrast, our method (\textbf{\ours}) accurately locates the specified region and performs the generation correctly.}
\vspace{-3mm}
    \label{fig:sota_cases}
\end{figure}

Unified Multimodal Models (UMMs)~\citep{nanobanana,gpt-4o,bagel,seedream2025seedream}, which unify generation and understanding across text and image modalities, have achieved notable success in precisely rendering photorealistic imagery from complex user prompts. Earlier advances in Large Language Models (LLMs) showed that purely text-based generation–understanding models can easily parse and reason over numerical coordinates in prompts~\citep{feng2023layoutgpt,omost}. This naturally motivates the question of whether UMMs, with their integrated visual generation capabilities, can not only understand spatial specifications in text but also translate that understanding into precise visual layouts. In practice, however, a notable gap emerges: even for simple instructions, current UMMs often fail to place or locate objects accurately at the specified coordinates on the canvas (see Fig.~\ref{fig:sota_cases}). This gap between textual spatial specifications and the rendered visual outcome limits their reliability in applications that demand strict spatial accuracy.

Researchers have worked hard to improve position control, yet existing methods still have clear limitations (see Tab.~\ref{tab:qualitative_comparison_single}). For instance, attention-masking approaches~\citep{zhou2025dreamrenderer,zhang2025eligen} provide only coarse, patch-level control. Adapter-based methods~\citep{zhang2024creatilayout,zhou20243dis,li2023gligen,wang2024instancediffusion} are difficult to integrate into a single unified model. Position-vocabulary methods, such as ReCo~\citep{yang2023reco}, rely on a large discrete vocabulary of position tokens, which increases computational and memory cost and prevents pixel-accurate specification of continuous coordinates.
In short, a persistent trade-off remains among accuracy, scalability, and architectural simplicity. \textit{A \textbf{lightweight, scalable, model-agnostic} solution that offers precise control remains an open challenge.}

\begin{figure*}[t]
  \centering
   \includegraphics[width=\linewidth]{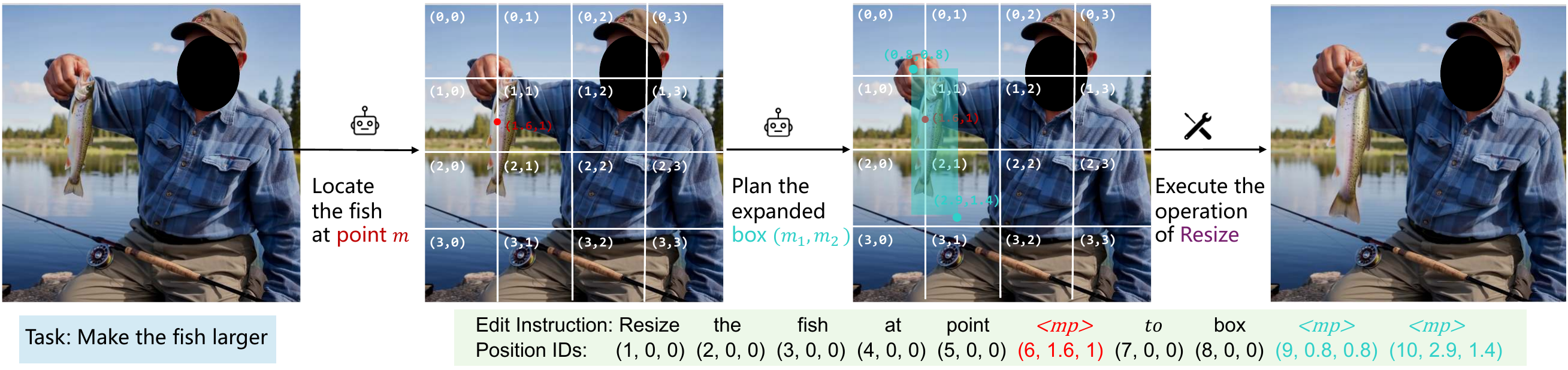}
   \vspace{-2mm}
\caption{\textbf{\ours and its agent.} (1) \textbf{\ours}, which uses special \texttt{<mp>} tokens linked to continuous 2D positional embeddings to localize targets (e.g., a fish) and to depict a guiding target box for the diffusion backbone; the \texttt{<mp>} positional embeddings are aligned with both the textual sequence position IDs and the 2D spatial coordinates of the corresponding pixels. (2) \textbf{\ours-Agent}, a Vision-Language Model (VLM), which decomposes complex tasks into sub-tasks, plans a solution pathway, and generates executable natural commands to drive the \ours model.}
\label{fig:ours-unified}
\vspace{-3mm}

    \label{fig:agent_mp}

\end{figure*}

Most modern generative models~\citep{bagel,mogao,nanobanana,gpt-4o,seedream2025seedream,labs2025flux,li2025onecat}, including UMMs, employ positional encoding schemes to represent locations in 2D visual space (Eqs.~\ref{eq:pe1}, \ref{eq:rope}). Motivated by this, we introduce \ours: a special textual token (\texttt{<mp>}) equipped with spatial positional encoding. The token’s word embedding conveys the intent to control position, while its positional embedding is engineered to encode the exact coordinates of a target point (see Fig.~\ref{fig:agent_mp}). This lightweight approach delivers pixel-level precision without architectural surgery, heavyweight vocabulary expansion, or coarse approximations.

By integrating this primitive into UMMs, we unlock a powerful compositional system for spatial control: a single \ours specifies a point; a pair defines a precise bounding box; and short sequences encode complex poses, trajectories, or object layouts. This transforms the UMM from a system that merely responds to ambiguous text into a precise surgical tool. It can now follow instructions augmented with \ours primitives to perform generation and editing with exact spatial awareness. Finally, to harness this compositional power and make it accessible for complex or casually phrased user requests, we introduce the \ours-Agent in Fig.~\ref{fig:agent_mp}. This VLM-based planner translates high-level user intent into a sequence of clear, explicit \ours-based instructions and acts as a crucial interpreter, converting ambiguous human language into precise, machine-executable commands that our enhanced UMM can execute.

Empirically, \ours achieves strong and consistent gains on diverse benchmarks: it improves mIoU on {COCO-MIG}~\cite{zhou2024migc} from 59.23\% to 77.29\% (\textbf{+30.49\%} relative) compared to the prior SOTA, boosts BAGEL’s overall score on {T2I-CoReBench}~\cite{t2icorebench} from 38.2 to 66.1 (\textbf{+73\%}), and raises {ImgEdit}~\cite{ye2025imgedit} overall from 3.42 to 3.94 (\textbf{+15.2\%}). Notably, the advantage grows with task difficulty (e.g., higher object counts), suggesting that explicit 2D positional grounding is a key inductive bias for robust spatial reasoning.

Beyond benchmarks, \ours shows two forms of scalability (Fig.~\ref{fig:teaser}): (i) reliable layout control for scenes with up to 30 objects while preserving visual fidelity; and (ii) simultaneous, coordinated editing of multiple objects of different types (\eg, replace, resize, move, and remove) without altering non-edited regions. Looking forward, \ours—as a stable, compositional primitive paired with a capable agent planner—forms a foundation for reliable, interactive generative systems that close the gap between numeric spatial intent and precise visual execution.

Our contributions are summarized as follows:
\begin{itemize}
    \item We propose \ours, a lightweight, model-agnostic, single-token interface that reuses the UMM's native positional encoding to achieve pixel-level spatial control---without architectural modifications, heavyweight vocabulary expansion, or coarse approximations.
    \item We show that \ours tokens are naturally compositional: a single token specifies a point, a pair defines a bounding box, and sequences encode complex layouts, or editing operations. Paired with a VLM-based planner (\ours-Agent), this enables an end-to-end system that translates high-level user intent into precise, spatially grounded generation and editing.
    \item Extensive experiments demonstrate that \ours establishes new state-of-the-art results on COCO-MIG, T2I-CoReBench, and ImgEdit, with gains that grow with task complexity, and scales reliably to scenes with up to 30 objects and multi-object coordinated editing.
\end{itemize}

\section{Related Work}
\label{sec:related}

\subsection{Precise Spatial Control in Image Generation}

Achieving precise spatial control in diffusion models has been a significant research focus. Early approaches relied on modifying the generation process with explicit guidance or masks. A prominent line of work involves \textbf{injecting spatial conditions via dedicated modules}. Methods like GLIGEN~\citep{li2023gligen,zhang2024creatilayout,zhou20253dis,zhou20243dis,zhou2024migc,wang2024instancediffusion,lee2024groundit} introduce attention mechanisms or feature injection modules directly into the UNet or DiT backbone. While effective, these methods require \textbf{substantial architectural modifications}, making them heavyweight and difficult to integrate into evolving model architectures. ReCo~\citep{yang2023reco} introduces 1,000 position-specific text tokens into the CLIP~\citep{clip} text encoder. However, it brings heavyweight vocabulary expansion and significant training overhead. Moreover, its discrete token design is \textbf{fundamentally imprecise and unscalable}, making it impossible to specify continuous coordinates, and it cannot extend to high-resolution generation without an exponential increase in its already-large token set.

\subsection{Agent-based Visual Generation}

Recent advancements in generative models~\citep{rombach2022high, bagel, nanobanana, seedream2025seedream,zhou2026refineanything,zhou2025bidedpo,ding2025muses,zhang2025layercraft,huang2024layerdiff,bader2025stitch,zhao2026zero,zhao2026luve,zhao2024wavelet,zhao2026ultrahr,zhao2025zero,lu2023tf,lu2024mace,lu2024robust,lu2025does,zhou2025dragflow} have enabled impressive capabilities. However, they consistently fail on tasks requiring high compositional fidelity and precise spatial control~\citep{geneval,dpgbench,huang2023t2i,wisebench,li2025gir,niu2025does,xie2025reconstruction,an2026genius,jin2025srum}. To mitigate these reasoning and compositional limitations, agent-based frameworks have emerged. \textbf{Prompt rewriting methods} employ an LLM to revise the user's initial prompt into a more detailed and suitable version for the generative model~\citep{wang2025promptenhancer,wang2025mint,jiang2025t2i,bagel,liao2025imagegen}, but this remains a text-only solution and is fundamentally incapable of fulfilling requests that require explicit spatial control. \textbf{Tool-calling methods} equip the agent with a "toolbox" of specialized and separated tools~\citep{wang2024genartist,fang2025got,li2026coco,yang2024mastering,wu2024self,zhang2024itercomp,wang2024divide} that are difficult to synergize and collaborate, leading to error propagation, poor controllability, and a failure to achieve stronger emergent functions, such as inserting a user-specific object into a given image. Our proposed \ours provides the \emph{explicit spatial control} that text-only solutions lack, while its \emph{native, unified design} avoids the synergy failures and error propagation inherent in fragmented toolboxes, providing a foundational component for a truly capable generative agent.

\section{\ours Method}
\label{sec:method}

Controlling object locations precisely and efficiently is central to generative AI. An ideal interface should (i) achieve pixel-level accuracy, (ii) use as few tokens as possible to avoid sequence-length and latency penalties, and (iii) plug into existing Unified Multimodal Models (UMMs) without architectural changes. Prior approaches do not meet all three goals simultaneously.

As shown in Fig.~\ref{fig:ours-unified}, we introduce \ours, a single-token interface for pixel-level spatial control. The core idea is to reuse the UMM's native image positional encoding to directly represent a continuous 2D coordinate with one special token \texttt{<mp>}. This design is minimal, token-efficient, and architecture-agnostic. We first revisit positional encoding in UMMs (\S\ref{subsec:pe}), then present the \ours token (\S\ref{subsec:mp}), and show how it enables spatial primitives (\S\ref{subsec:spatial_primitives}).

\begin{table}[t]
\centering
\scriptsize
\caption{
\textbf{Comparison of position-control methods.}
\ours uniquely achieves pixel-level precision, single-token efficiency, and native UMM compatibility. We evaluate precision (Prec.), vocab token additions, and UMM compatibility.
}
\vspace{-3mm}
\label{tab:qualitative_comparison_single}
{
\begin{tabular}{l l c c }
\toprule
\textbf{Method} & \textbf{Prec.} & \textbf{Add Tokens} & \textbf{UMM} \\
\midrule
Text Prompt~\cite{nanobanana,gpt-4o,seedream2025seedream} & \emph{roughly} & 0 & \CheckmarkBold  \\
\addlinespace
Attn Mask~\citep{zhou2025dreamrenderer,zhang2025eligen} & \emph{patch} & 0 & \XSolidBrush  \\
\addlinespace
Adapter~\citep{zhou20243dis,zhang2024creatilayout} & \emph{pixel} & 0 & \XSolidBrush \\
Position Vocabulary~\citep{yang2023reco} & $\approx$ \emph{pixel} & N & \XSolidBrush  \\
\midrule
\ours & \emph{pixel} & 1 & \CheckmarkBold \\
\bottomrule
\vspace{-3mm}
\end{tabular}}
\end{table}

\subsection{Positional Encoding in UMMs}
\label{subsec:pe}

UMMs reason over tokens embedded in a 3D index space: a sequence axis for order and two spatial axes (height, width) for layout. Two families of positional encoding are common: a hybrid scheme using 1D RoPE~\cite{su2024roformer} for sequence and 2D Sinusoidal Positional Embedding~\citep{attention} for space, and a unified 3D RoPE that applies rotary embeddings across all axes. The main difference is whether sequence and spatial positions are handled by separate or shared mechanisms.

\noindent\textbf{2D Sinusoidal PE.}
Let $(u,v)$ denote the spatial coordinates of an image token. The 2D sinusoidal positional embedding maps $(u,v)$ to a $d$-dimensional vector by splitting the dimensions into two halves: the first $d/2$ encode the row $u$, and the second $d/2$ encode the column $v$,
\begin{equation}
\label{eq:pe1}
\begin{aligned}
&\text{PE}_u(i)=\sin\!\left(\frac{u}{K^{4i/d}}\right), \quad\text{PE}_u(i+d/4)=\cos\!\left(\frac{u}{K^{4i/d}}\right),
\end{aligned}
\end{equation}
where $i \in \{0, 1, \dots, d/4-1\}$ and $K$ is a base (e.g., 10000). The second half encodes $v$ in the same way, and the final embedding is the concatenation:
\begin{equation}
\label{eq:pe2}
\text{PE} = [\text{PE}_u, \text{PE}_v].
\end{equation}
The visual token embedding is augmented by adding PE.

\noindent\textbf{3D RoPE.}
In self-attention~\citep{vaswani2017attention}, Rotary Position Embedding (RoPE)~\citep{su2024roformer} applies position-dependent rotations. For UMMs, the embedding dimension $d$ is partitioned into three blocks for the sequence position, height, and width ($d = d_1 + d_2 + d_3$). Independent rotations are applied to each block with respect to the corresponding index $m$:
\begin{equation}
\label{eq:rope}
    \mathcal{R}(m) = \begin{pmatrix} \cos m\theta & -\sin m\theta \\ \sin m\theta & \cos m\theta \end{pmatrix}.
\end{equation}
For a visual token at $(u,v)$, rotations $\mathcal{R}(u)$ and $\mathcal{R}(v)$ are applied to the spatial chunks using a list of angles $\theta$. See BAGEL~\citep{bagel} and Mogao~\citep{mogao} for details.

\subsection{\ours}
\label{subsec:mp}

UMMs already encode spatial relations via their positional systems. We ask: how can we directly leverage this native capability to represent a specific 2D location with maximal precision and minimal overhead?

We define a single special token, \texttt{<mp>}, that serves as a pointer to a continuous 2D coordinate $(u,v)$. When \texttt{<mp>} is used, the UMM applies its existing spatial positional encoding (e.g., 2D Sinusoidal PE or 3D RoPE) to $(u,v)$ as if it were an image token\footnote{BAGEL uses 2D Sinusoidal PE, and all results in the paper use \ours with the same PE. We also implemented \ours on an in-house UMM with a 3D RoPE variant.}, as illustrated in Fig.~\ref{fig:agent_mp}. Concretely, the embedding of \ours at $(u,v)$ is
\begin{equation}
\label{eq:metapoint_embedding}
\mathrm{X}_{u,v} = \mathrm{X}_{\texttt{<mp>}} + [\text{PE}_u, \text{PE}_v],
\end{equation}
where $\mathrm{X}_{\texttt{<mp>}}$ is the learned vocabulary embedding for the special token.

Crucially, we treat $(u,v)$ as continuous coordinates. Although positional encodings are often applied to discrete patch indices, the formulas in \eqref{eq:pe1}, \eqref{eq:pe2}, and \eqref{eq:rope} are differentiable and accept floating-point inputs. Supplying continuous $(u,v)$ breaks free from the patch grid and enables pixel-level control independent of the generation resolution.

Intuitively, \ours models a virtual ``click'' on the canvas: the click extracts the exact location and compresses it into one token. The design is notable for its simplicity and effectiveness: (i) one token per point, (ii) no architectural changes, and (iii) direct reuse of the model's spatial reasoning. \ours attains both precision and efficiency without compromise.

\subsection{Versatile Spatial Control via \ours}
\label{subsec:spatial_primitives}

\begin{figure*}[t]
  \centering
  \includegraphics[width=\linewidth]{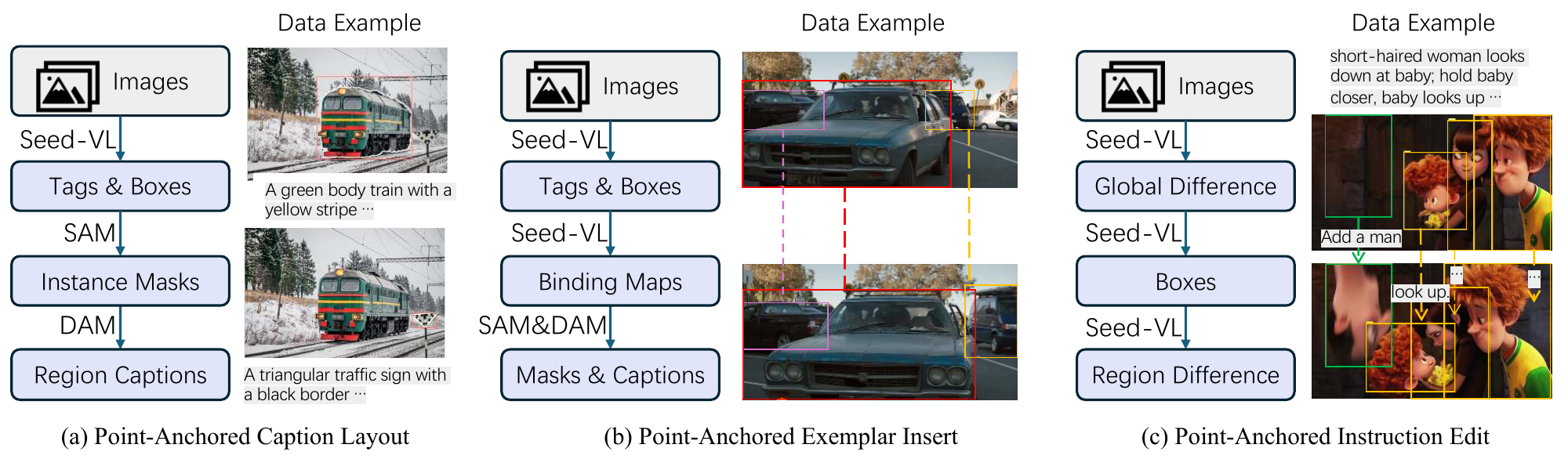}
\caption{\textbf{Point-Anchored Data Pipeline.}
\textbf{(a) Caption Layout:} Single image; tags/captions/boxes/masks; bind text captions to \ours{}s.
\textbf{(b) Exemplar Insert:} Paired frames; boxes and binding maps for the same object between two frames; bind visual exemplar to \ours{}.
\textbf{(c) Instruction Edit:} Paired frames; global diffs $\rightarrow$ regional edit instruction; bind \ours{}s to multi-region edits.}


  \label{fig:data_engine}
  
\end{figure*}

\begin{table}[!t]
\centering
\scriptsize
\caption{
    \textbf{Statistics of the training datasets.}
    We mix some in-house datasets (first block) with our newly constructed datasets (second block) and report their sampling ratios.
    Acronyms (PACL, PAEI, PAIE) are defined in Sec.~\ref{sec:data_engine}.
}
\vspace{-3mm}

\label{tab:dataset_statistics}
\begin{tabular}{l c c c}
\toprule
\textbf{Dataset} & \textbf{\# Samples} & \textbf{Ratio} & \textbf{Supporting Task} \\
\midrule
T2I Data & 1M & 3.8\% & Maintain t2i generation \\
\addlinespace
OCR Data & 50K & 0.2\% & Text edit\\
\midrule
PACL & 3M & 31\% & Caption-driven layout \\
\addlinespace
PAEI & 3M & 35\% & Exemplar-based insertion \\
\addlinespace
PAIE & 2M & 30\% & Instruction-driven edit \\
\bottomrule
\vspace{-3mm}
\end{tabular}
\end{table}

\ours enables versatile spatial primitives (Fig.~\ref{fig:teaser}), and this section shows how \ours can be composed to achieve spatial control in different tasks, from generation to editing, and how they integrate with a planner (e.g., a VLM) for instruction-driven workflows and self-reflective autonomous correction:

\begin{itemize}
    \item \textbf{Generation: layout and count.}
    A single token \texttt{<mp>} can specify where an object should be generated.
    Two tokens $[\texttt{mp}_{\text{tl}}, \texttt{mp}_{\text{br}}]$ define a bounding box, allowing both position and size control.
    A sequence of $N$ tokens $[\texttt{mp}_1, \dots, \texttt{mp}_N]$ specifies the layout of $N$ objects.

    \item \textbf{Object editing.}
    Edits can be conditioned on a \emph{source} box alone or on a \emph{source--target} pair.
    In the first case, a token sequence such as $[\texttt{mp}_{\text{src1}}, \texttt{mp}_{\text{src2}}]$ specifies the box of the object to be modified, while the desired change (e.g., recolor, replace) is described in natural language.
    In the second case, both a source and a target box are provided, e.g.,
    $[\texttt{mp}_{\text{src1}}, \texttt{mp}_{\text{src2}}] \rightarrow [\texttt{mp}_{\text{tgt1}}, \texttt{mp}_{\text{tgt2}}]$,
    instructing the model to transform the source object to fit the target region---moving, resizing, or modifying pose.
    This flexible mechanism supports a wide range of spatial edits, including insertion, replacement, moving, and resizing.

    \item \textbf{Integration with a VLM Planner.}
    A VLM~\cite{guo2025seedvl} bridges user intent and low-level \ours primitives. As shown in Fig.~\ref{fig:agent_mp}, the VLM (1) \textbf{perceives} the image and localizes the fish as $\texttt{mp}_{\text{src}}$, (2) \textbf{reasons} about the intent to plan a target box $[\texttt{mp}_{\text{t1}}, \texttt{mp}_{\text{t2}}]$, and (3) \textbf{generates} the natural language instruction based on \ours primitives.

    \item \textbf{Self-reflection and Autonomic Correction.}
    The precision of \ours-based editing enables a VLM Planner not only to generate and edit content, but also to \emph{self-reflect} on its own outputs.
    After generating an image, the VLM can re-analyze the result, localize specific spatial errors (e.g., missing objects, incorrect size or count), and solve each problem with exact bounding boxes via \ours tokens.
    It can then issue corrective edit instructions, such as ``add a cubic plaster'' or ``remove the sphere'' with precise coordinates (Fig.~\ref{fig:teaser}).
    This closed-loop process---\emph{generate $\rightarrow$ reflect $\rightarrow$ execute}---enables autonomous improvement without human intervention.
\end{itemize}

\section{Data Pipeline}
\label{sec:data_engine}

Video provides rich supervision via temporal consistency and variation. Harnessing this principle, our data engine automatically constructs three distinct datasets to enable precise spatial control, \eg, layout, insertion, and editing. For brevity, we denote these datasets as \textbf{PACL} (Point-Anchored Caption Layout), \textbf{PAEI} (Point-Anchored Exemplar Insert), and \textbf{PAIE} (Point-Anchored Instruction Edit) (see Fig.~\ref{fig:data_engine} for an overview).

\noindent\textbf{PACL.}
We generate layout data with dense supervision (Fig.~\ref{fig:data_engine}\emph{a}): Seed-VL~\citep{guo2025seedvl} yields precise tags and tight boxes, SAM~\citep{sa1b} provides pixel masks, and DAM~\citep{lian2025describe} produces region captions. Each image is annotated with tags, boxes, masks, and captions anchored to \ours.

\noindent\textbf{PAEI.}
For visual exemplar insertion (Fig.~\ref{fig:data_engine}\emph{b}), we sample frame pairs from videos, using Seed-VL to detect objects and establish correspondences. For training, one frame serves as the ground truth (GT) frame. Objects from the other frame, specified by their bounding box in the GT frame, become the visual exemplar (grounded image condition). The model learns to insert this visual exemplar at a location specified by the MetaPoint.

\noindent\textbf{PAIE.}
For editing (Fig.~\ref{fig:data_engine}\emph{c}), we pair frames from videos, detect global and region-level changes with Seed-VL, and auto-generate concise edit instructions (\eg, add, move, resize, remove) bound to \ours. This trains precise, localized edits while preserving the background.

\section{Experiments}

\begin{table*}[!t]
\centering
\scriptsize
\caption{Quantitative comparison on \textbf{COCO-MIG} for multi-instance generation tasks. \textbf{Instance Success Rate (\%) $\uparrow$} and \textbf{mIoU $\uparrow$} metrics across different object count levels ($L_2$ to $L_6$).}
\vspace{-3mm}
\label{tab:cocomig}
\definecolor{sotablue}{HTML}{2E86C1}

\resizebox{\textwidth}{!}{
\begin{tabular}{l cccccc cccccc}
\toprule
\textbf{Method} & \multicolumn{6}{c}{\textbf{Instance Success Rate(\%) $\uparrow$}} & \multicolumn{6}{c}{\textbf{mIoU $\uparrow$}} \\
\cmidrule(lr){2-7} \cmidrule(lr){8-13}
 & Avg & $L_2$ & $L_3$ & $L_4$ & $L_5$ & $L_6$ & Avg & $L_2$ & $L_3$ & $L_4$ & $L_5$ & $L_6$ \\
\midrule
LAMIC~\cite{chen2025lamic} & 13.56 & 28.12 & 19.17 & 13.75 & 9.00 & 9.58 & 21.17 & 31.67 & 25.79 & 20.68 & 18.08 & 18.25 \\
GrounDiT~\cite{lee2024groundit} & 22.91 & 36.56 & 31.25 & 22.97 & 17.75 & 18.44 & 29.72 & 37.41 & 35.30 & 30.13 & 26.50 & 26.79 \\
GLIGEN~\cite{li2023gligen} & 29.56 & 41.88 & 31.67 & 27.19 & 27.38 & 27.81 & 27.44 & 37.35 & 29.17 & 25.31 & 26.42 & 25.56 \\
MS-Diffusion~\cite{wang2025msdiffusion} & 28.22 & 37.81 & 33.12 & 28.12 & 25.75 & 24.69 & 34.69 & 41.15 & 36.38 & 34.57 & 32.36 & 33.70 \\
CreatiLayout~\cite{zhang2025creatilayout} & 54.69 & 67.19 & 63.33 & 56.09 & 50.25 & 48.96 & 48.96 & 56.32 & 55.38 & 49.42 & 46.22 & 45.28 \\
InstanceDiffusion~\cite{wang2024instancediffusion} & 60.28 & 71.25 & 61.67 & 59.38 & 57.00 & 59.27 & 54.79 & 65.76 & 57.21 & 53.33 & 51.43 & 53.72 \\
ReCo~\cite{yang2023reco} & 56.90 & 65.50 & 56.10 & 56.30 & 52.40 & 58.30  & 47.60 & 55.70 & 46.70 & 47.20 & 43.30 & 48.80 	 \\
EliGen~\cite{zhang2025eligen} & 64.12 & 69.69 & 72.50 & 66.56 & 61.62 & 58.54 & 59.23 & 64.61 & 66.10 & 61.59 & 56.74 & 54.50 \\
MIGC~\cite{zhou2024migc} & 66.44 & 74.06 & 67.29 & 67.03 & 63.25 & 65.73 & 56.96 & 63.84 & 57.60 & 56.95 & 54.01 & 56.82 \\
\midrule

\textbf{BAGEL+\ours} & \textbf{84.72} & \textbf{84.52} & \textbf{84.31} & \textbf{86.66} & \textbf{83.29} & \textbf{84.85} & \textbf{77.29} & \textbf{76.72} & \textbf{76.60} & \textbf{79.32} & \textbf{76.22} & \textbf{77.34} \\
\textcolor{sotablue}{\vs SOTA} & \textcolor{sotablue}{+18.28} & \textcolor{sotablue}{+10.46} & \textcolor{sotablue}{+11.81} & \textcolor{sotablue}{+19.63} & \textcolor{sotablue}{+20.04} & \textcolor{sotablue}{+19.12} & \textcolor{sotablue}{+18.06} & \textcolor{sotablue}{+10.96} & \textcolor{sotablue}{+10.50} & \textcolor{sotablue}{+17.73} & \textcolor{sotablue}{+19.48} & \textcolor{sotablue}{+20.52} \\
\bottomrule
\end{tabular}%
}

\vspace{3mm}
\centering
\includegraphics[width=0.95\linewidth]{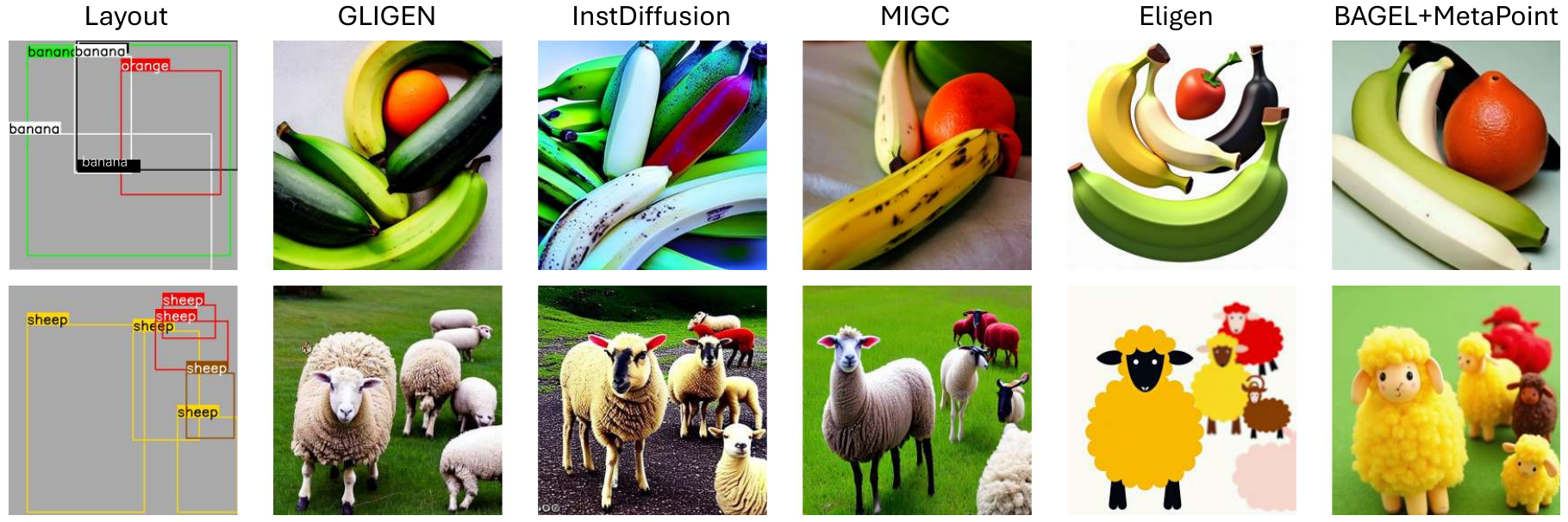}
\vspace{-2mm}
\captionof{figure}{
\textbf{Qualitative Results on COCO-MIG benchmark.} Each instance is assigned a location and color, shown by its bounding box.
}
\label{fig:cocomig_vis}
\vspace{-3mm}
\end{table*}

\subsection{Experimental Setup}
\noindent\textbf{Implementation Details.}
We train our model, \ours, on the datasets detailed in~\cref{tab:dataset_statistics} for 10K steps, starting from BAGEL~\citep{bagel}. The training is conducted on 256 H20 GPUs and takes approximately two days to complete. For a fair comparison during inference, our method strictly reuses the original configuration of BAGEL, including its resolution, sampler, steps, and CFG scale.

\noindent\textbf{Evaluation.}
We evaluate our method on three benchmarks that cover both generative and editing tasks.
\begin{itemize}
    \item \textbf{COCO-MIG}~\cite{zhou2024migc}: Layout-to-image benchmark evaluating a model’s ability to control object placement and attribute bindings in multi-instance scenes.
    \item \textbf{T2I-CoReBench}~\cite{t2icorebench}: Text-to-image benchmark for assessing a model’s composition and reasoning capabilities.
    \item \textbf{ImgEdit}~\cite{ye2025imgedit}: Image editing benchmark that measures performance across nine instruction-driven categories.
\end{itemize}


\begin{table*}[!t]
\centering
\caption{\textbf{Main results on \textsc{T2I-CoReBench}} assessing both \textit{composition} and \textit{reasoning} capabilities.}
\vspace{-3mm}
\definecolor{sotablue}{HTML}{2E86C1}
\definecolor{meangray}{HTML}{E2E2E2}
\definecolor{overallbeige}{HTML}{FDF3E3} 

\resizebox{\hsize}{!}{
\begin{tabular}{lccccccccccccccc}
\toprule
& \multicolumn{5}{c}{{\textbf{Composition}}} & \multicolumn{9}{c}{{\textbf{Reasoning}}} & \\ \cmidrule(lr){2-15}
\multirow{-2}{*}{\textbf{Model}} & MI & MA & MR & TR & \cellcolor{meangray}Mean & LR & BR & HR & PR & GR & AR & CR & RR & \cellcolor{meangray}Mean & \multirow{-2}{*}{\textbf{Overall}} \\ \midrule

\multicolumn{16}{c}{{\textbf{Closed-Source Models}}} \\ \midrule
GPT-4o-Image~\citep{gpt-4o} & 84.1 & 75.9 & 72.7 & 86.4 & \cellcolor{meangray}79.8 & 59.0 & 54.8 & 65.6 & 87.3 & 76.5 & 82.0 & 70.9 & 56.1 & \cellcolor{meangray}69.0 & \cellcolor{overallbeige}72.6 \\ 
Seedream 4.0~\citep{seedream2025seedream} & 91.5 & 84.5 & 75.0 & 93.6 & \cellcolor{meangray}86.1 & 76.3 & 54.1 & 60.7 & 85.8 & 85.9 & 77.1 & 71.6 & 47.9 & \cellcolor{meangray}69.9 & \cellcolor{overallbeige}75.3 \\
Nano Banana~\citep{nanobanana} & 85.7 & 77.9 & 72.6 & 86.3 & \cellcolor{meangray}80.6 & 64.5 & 64.9 & 67.1 & 85.2 & 84.1 & 83.1 & 71.3 & 68.7 & \cellcolor{meangray}73.6 & \cellcolor{overallbeige}75.9 \\
Imagen 4 Ultra~\citep{imagen4} & 90.0 & 80.0 & 73.2 & 86.2 & \cellcolor{meangray}82.4 & 63.6 & 62.4 & 66.1 & 88.5 & 82.8 & 83.0 & 76.3 & 60.7 & \cellcolor{meangray}72.9 & \cellcolor{overallbeige}76.1 \\ 

\midrule

\multicolumn{16}{c}{{\textbf{Open-Source Models}}} \\ \midrule

Janus-Pro-7B~\citep{janus} & 54.4 & 59.3 & 40.9 & 7.5 & \cellcolor{meangray}40.5 & 19.8 & 20.9 & 34.6 & 22.4 & 11.5 & 30.4 & 8.7 & 9.8 & \cellcolor{meangray}19.8 & \cellcolor{overallbeige}26.7 \\
SD-3.5-Large~\citep{esser2024scaling} & 57.5 & 60.0 & 32.9 & 15.6 & \cellcolor{meangray}41.5 & 22.5 & 22.4 & 34.2 & 52.5 & 35.5 & 53.0 & 42.3 & 25.2 & \cellcolor{meangray}35.9 & \cellcolor{overallbeige}37.8 \\
BAGEL~\citep{bagel} & 64.9 & 65.2 & 45.8 & 9.7 & \cellcolor{meangray}46.4 & 23.4 & 21.9 & 33.0 & 51.6 & 31.2 & 50.4 & 32.4 & 29.3 & \cellcolor{meangray}34.1 & \cellcolor{overallbeige}38.2 \\
OmniGen2-7B~\citep{wu2025omnigen2explorationadvancedmultimodal} & 67.9 & 64.1 & 48.3 & 19.2 & \cellcolor{meangray}49.9 & 24.7 & 23.2 & 43.3 & 63.1 & 46.1 & 54.2 & 36.5 & 24.1 & \cellcolor{meangray}39.4 & \cellcolor{overallbeige}42.9 \\
HiDream-I1~\citep{hidreami1technicalreport} & 62.5 & 62.0 & 42.9 & 33.9 & \cellcolor{meangray}50.3 & 34.2 & 24.5 & 40.9 & 53.2 & 34.2 & 50.3 & 46.1 & 31.7 & \cellcolor{meangray}39.4 & \cellcolor{overallbeige}43.0 \\
FLUX.1-Krea-dev~\citep{labs2025flux} & 70.7 & 71.1 & 53.2 & 28.9 & \cellcolor{meangray}56.0 & 30.3 & 26.1 & 44.5 & 70.6 & 50.5 & 57.5 & 46.3 & 28.7 & \cellcolor{meangray}44.3 & \cellcolor{overallbeige}48.2 \\
Qwen-Image~\citep{wu2025qwen} & \textbf{81.4} & \textbf{79.6} & \textbf{65.6} & \textbf{85.5} & \cellcolor{meangray}{\textbf{78.0}} & 41.1 & 32.2 & 48.2 & 75.1 & 56.5 & 53.3 & \textbf{61.9} & 26.4 & \cellcolor{meangray}49.3 & \cellcolor{overallbeige}58.9 \\ 
\midrule
\textbf{BAGEL + \ours} & 79.4 & 72.7 & 54.7 & 48.4 & \cellcolor{meangray}{66.3}\ & \textbf{73.3} & \textbf{53.4} & \textbf{62.2} & \textbf{80.5} & \textbf{79.1} & \textbf{67.0} & 56.6 & \textbf{55.9} & \cellcolor{meangray}{\textbf{66.0}} & \cellcolor{overallbeige}{\textbf{66.1}} \\
\textcolor{sotablue}{\vs BAGEL} & \textcolor{sotablue}{+14.5} & \textcolor{sotablue}{+7.5} & \textcolor{sotablue}{+8.9} & \textcolor{sotablue}{+38.7} & \cellcolor{meangray}{\textcolor{sotablue}{+19.9}} & \textcolor{sotablue}{+49.9} & \textcolor{sotablue}{+31.5} & \textcolor{sotablue}{+29.2} & \textcolor{sotablue}{+28.9} & \textcolor{sotablue}{+47.9} & \textcolor{sotablue}{+16.6} & \textcolor{sotablue}{+24.2} & \textcolor{sotablue}{+26.6} & \cellcolor{meangray}{\textcolor{sotablue}{+31.9}} & \cellcolor{overallbeige}{\textcolor{sotablue}{+27.9}} \\ 
\bottomrule
\end{tabular}
}
\label{table:t2icore_with_ours_final}

\vspace{3mm}
\centering
\includegraphics[width=0.95\linewidth]{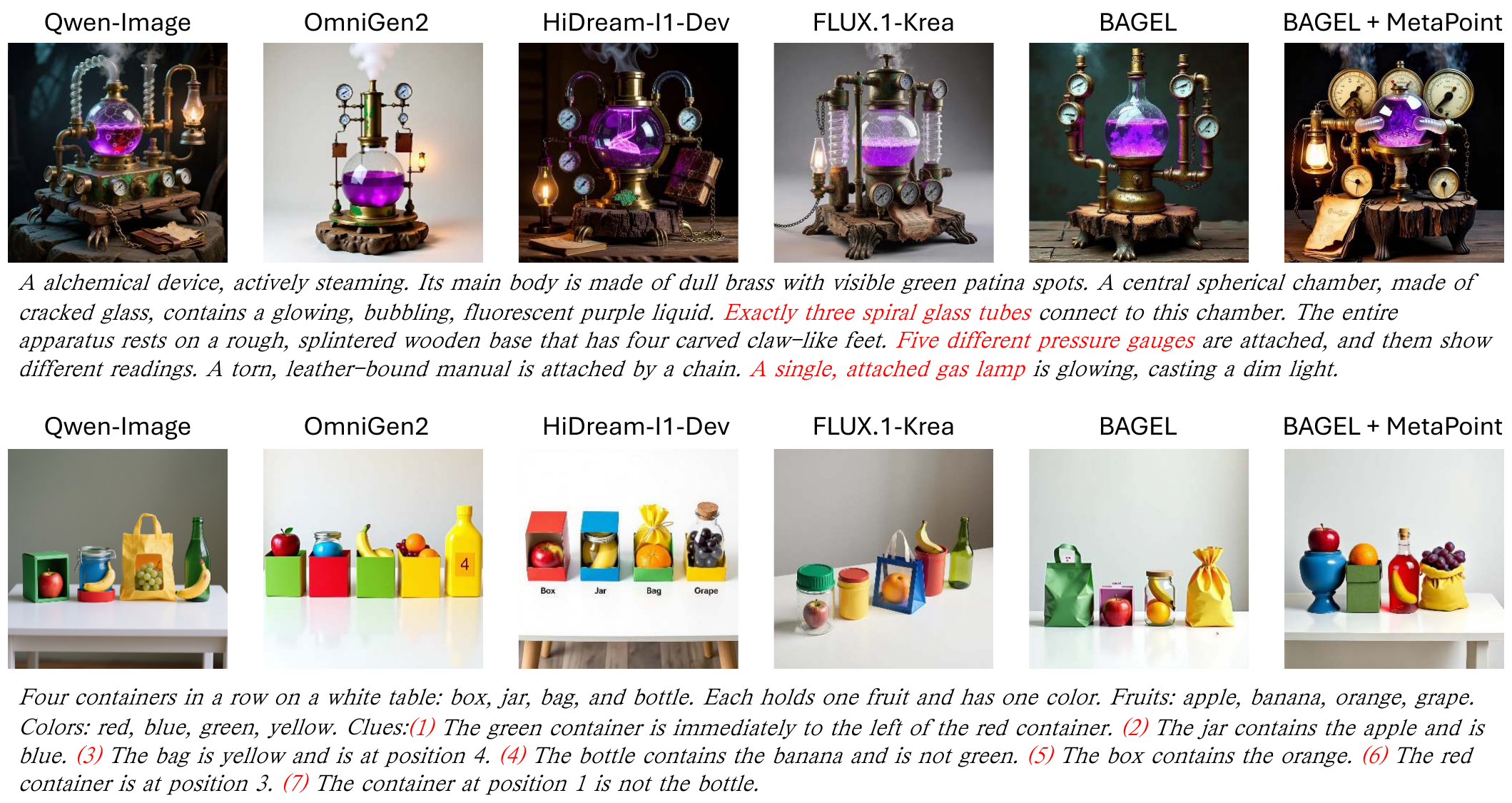}
\vspace{-2mm}
\captionof{figure}{
\textbf{Qualitative Results on T2I-CoReBench.}
}
\label{fig:t2i_vis}
\vspace{-3mm}
\end{table*}

For all benchmarks, we adapt a VLM to translate all tasks into a new instruction with \texttt{<mp>}\footnote{The utilized system prompts are given in supplementary materials.} and then follow the original evaluation protocols. We report {Instance Success Rate (\%) $\uparrow$} (measuring attribute accuracy) and {mIoU $\uparrow$} (measuring layout accuracy) for COCO-MIG, use the official Gemini 2.5 Flash~\citep{comanici2025gemini} for scoring on T2I-CoReBench, and employ GPT-4.1 to score instruction alignment, visual fidelity, and realism (on a 1--5 scale) for ImgEdit.

\subsection{Quantitative and Qualitative Results}

\noindent\textbf{COCO-MIG.} As shown in~\cref{tab:cocomig}, our method substantially outperforms the prior SOTA on COCO-MIG, raising the average Instance Success Rate from 66.44\% to 84.72\% (\textbf{+27.51\%} relative) and mIoU from 59.23\% to 77.29\% (\textbf{+30.49\%} relative). The advantage grows with task difficulty: as object count increases (L2$\rightarrow$L6), Instance Success Rate gains expand from {+10.46} (L2) to over {+19} (L4--L6), and mIoU gains rise from {+10.96} (L2) to {+20.52} (L6). Performance is stable across levels (L2--L6), with Instance Success Rate at $\approx$84--87\% and mIoU at $\approx$76--79\%, indicating the benchmark’s complexity is well within our model’s capabilities. Qualitative results (Fig.~\ref{fig:cocomig_vis}) show more natural, coherent scenes that satisfy all constraints. While COCO-MIG is limited to 6 objects, \ours scales well beyond the benchmark, controlling up to 30 objects with high fidelity (\cref{fig:teaser}), demonstrating a capability that exceeds current methods.

\begin{table*}[!t]
\centering
\scriptsize
\caption{\textbf{Comparison results on ImgEdit.}}

\vspace{-3mm}
\label{tab:tab_ImgEdit_modified}
\definecolor{sotablue}{HTML}{2E86C1}

\resizebox{\textwidth}{!}{
\begin{tabular}{lcccccccccc}
\toprule
\textbf{Model} & Add & Adjust & Extract & Replace & Remove & Background & Style & Hybrid & Action & Overall$\uparrow$ \\
\midrule
UltraEdit~\citep{zhao2024ultraedit} & 3.44 & 2.81 & 2.13 & 2.96 & 1.45 & 2.83 & 3.76 & 1.91 & 2.98 & 2.70 \\
ICEdit~\citep{zhang2025context} & 3.58 & 3.39 & 1.73 & 3.15 & 2.93 & 3.08 & 3.84 & 2.04 & 3.68 & 3.05 \\
Step1X-Edit~\citep{liu2025step1x} & 3.88 & 3.14 & 1.76 & 3.40 & 2.41 & 3.16 & 4.63 & 2.64 & 2.52 & 3.06 \\
UniWorld-V1~\citep{lin2025uniworld} & 3.82 & 3.64 & 2.27 & 3.47 & 3.24 & 2.99 & 4.21 & 2.96 & 2.74 & 3.26 \\
BAGEL~\citep{bagel} & 3.81 & 3.59 & 1.58 & 3.85 & 3.16 & 3.39 & 4.51 & 2.67 & 4.25 & 3.42 \\
OmniGen2~\citep{wu2025omnigen2explorationadvancedmultimodal} & 3.57 & 3.06 & 1.77 & 3.74 & 3.20 & 3.57 & 4.81 & 2.52 & 4.68 & 3.44 \\
FLUX-Kontext~\citep{labs2025flux} & 3.83 & 3.65 & 2.27 & 4.45 & 3.17 & 3.98 & 4.55 & 3.35 & 4.29 & 3.71 \\
GPT-4o-Image & \textbf{4.61} & \textbf{4.33} & 2.90 & 4.35 & 3.66 & \textbf{4.57} & \textbf{4.93} & \textbf{3.96} & \textbf{4.89} & 4.20 \\
Qwen-Image & 4.38 & 4.16 & \textbf{3.43} & \textbf{4.66} & 4.14 & 4.38 & 4.81 & 3.82 & 4.69 & \textbf{4.27}  \\
\midrule
\textbf{BAGEL + \ours} & 4.26 & 4.14 & 2.24 & 4.32 & \textbf{4.20} & 3.82 & 4.82 & 3.07 & 4.22 & 3.94 \\
\textcolor{sotablue}{\vs BAGEL} & \textcolor{sotablue}{+0.45} & \textcolor{sotablue}{+0.55} & \textcolor{sotablue}{+0.66} & \textcolor{sotablue}{+0.47} & \textcolor{sotablue}{+1.04} & \textcolor{sotablue}{+0.43} & \textcolor{sotablue}{+0.31} & \textcolor{sotablue}{+0.40} & \textcolor{sotablue}{-0.03} & \textcolor{sotablue}{+0.52} \\
\bottomrule
\end{tabular}
}

\vspace{3mm}
\centering
\includegraphics[width=0.95\linewidth]{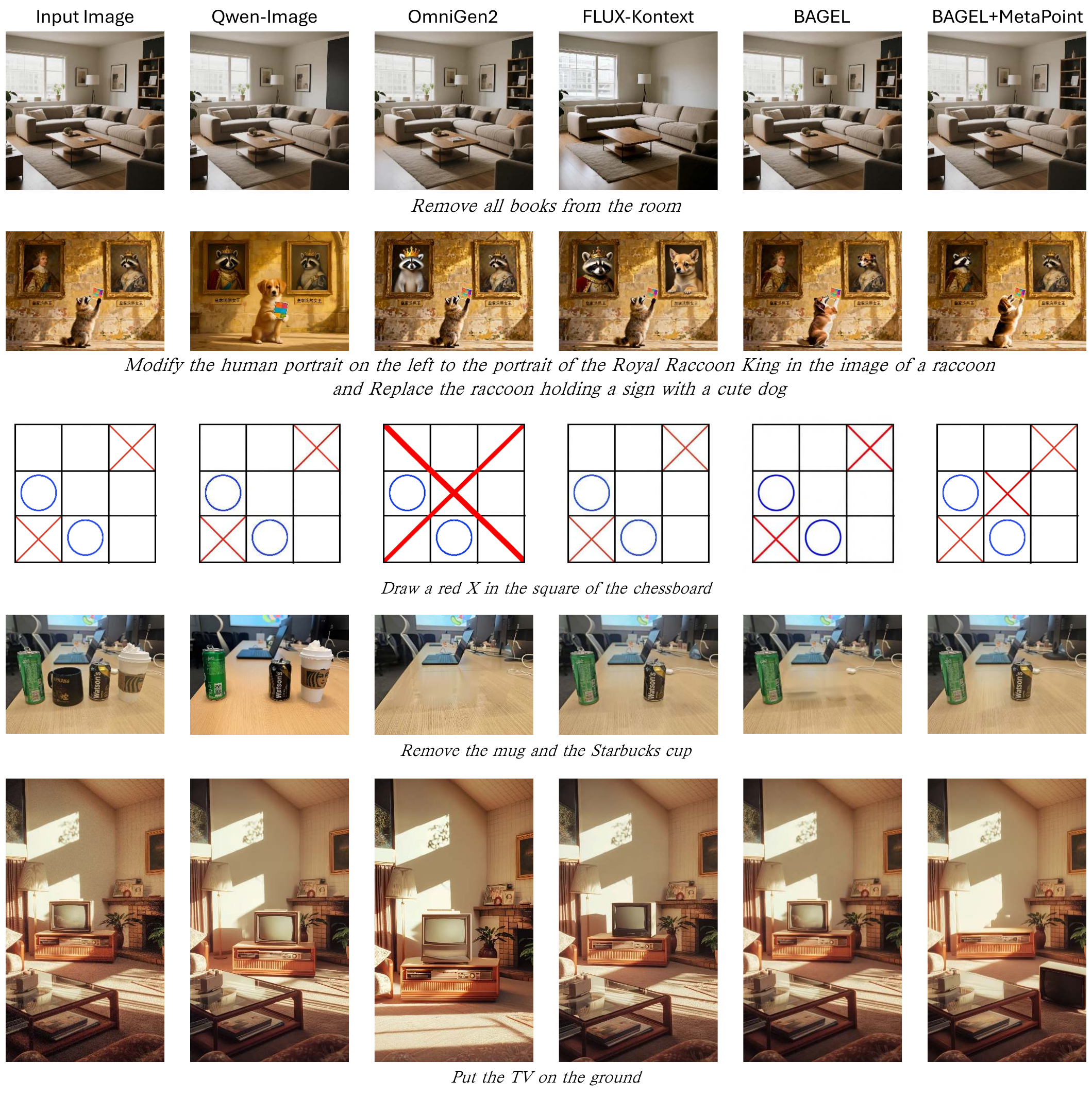}
\vspace{-3mm}
\captionof{figure}{
\textbf{Qualitative comparison on complex editing tasks.}
}
\label{fig:edit_vis}
\vspace{-3mm}
\end{table*}

\noindent\textbf{T2I-CoReBench.}
Our method boosts the BAGEL baseline's overall score from 38.2 to 66.1, a \textbf{73\%} relative improvement that establishes a new state-of-the-art for open-source models (Table~\ref{table:t2icore_with_ours_final}). This improvement is driven by substantial gains in both Composition (mean score {+19.9}) and particularly in Reasoning (mean score {+31.9}). Notably, our method delivers its largest gains on sub-tasks where the baseline was particularly weak, such as Logical Reasoning (LR, {+49.9}), Geometric Reasoning (GR, {+47.9}), and Text Rendering (TR, {+38.7}). As shown in Fig.~\ref{fig:t2i_vis}, these quantitative leaps are corroborated by qualitative results, where our model successfully handles nuanced prompts that challenge the baseline.

\noindent{\textbf{ImgEdit.}}
As shown in \cref{tab:tab_ImgEdit_modified}, \ours significantly enhances BAGEL, boosting the Overall score: \textbf{3.42} $\rightarrow$ \textbf{3.94} (\textbf{+15.2\%} relative). While Qwen-Image achieves a higher Overall score (4.27), our method excels in specific tasks like \emph{Remove} ({3.16} $\rightarrow$ {4.20}), where it achieves the best score.
This performance is achieved using BAGEL with minimal edit-specific training, guided by a VLM agent that translates user requests into structured \ours-based operations (\eg, move, resize). 
Beyond benchmarks, as shown in Fig.~\ref{fig:edit_vis}, \ours performs precise localized modifications while preserving the background, unlike text-only baselines that often cause grounding errors and unintended global changes.

\subsection{Ablation of Positional Encoding}
We ablate positional encoding while holding data, compute, and sampling constant. \textbf{BAGEL + Text} supplies natural-language coordinates (normalized to [0, 1000]) to the UMM and relies on the model to infer geometry from tokens; \textbf{BAGEL + \ours} instead injects spatially aligned 2D positional embeddings directly into the DiT branch.

As reported in~\cref{tab:ablate_mp}, replacing text coordinates with \ours yields a large improvement on COCO-MIG: average Instance Success Rate rises from 61.84\% to 84.72\% ({+22.88}) and average mIoU from 52.48\% to 77.29\% ({+24.81}). Notably, the text-based variant exhibits high variance across complexity levels (ISR ranges from 50.00 at $L_2$ to 66.77 at $L_6$), whereas \ours remains consistently strong ($\approx$83--87\% ISR across all levels), indicating that explicit 2D positional encoding generalizes robustly regardless of scene complexity.
\begin{table*}[!t]
\centering
\scriptsize
\caption{\textbf{Ablation of \ours. (Text \vs \ours)}}

\vspace{-3mm}
\label{tab:ablate_mp}
\definecolor{sotablue}{HTML}{2E86C1} 

\resizebox{\textwidth}{!}{
\begin{tabular}{l cccccc cccccc}
\toprule
\textbf{Method} & \multicolumn{6}{c}{\textbf{Instance Success Rate(\%) $\uparrow$}} & \multicolumn{6}{c}{\textbf{mIoU $\uparrow$}} \\
\cmidrule(lr){2-7} \cmidrule(lr){8-13}
 & Avg & $L_2$ & $L_3$ & $L_4$ & $L_5$ & $L_6$ & Avg & $L_2$ & $L_3$ & $L_4$ & $L_5$ & $L_6$ \\
\midrule

BAGEL+Text & 61.84 & 50.00 & 61.44 & 60.30 & 62.14 & 66.77 & 52.48 & 47.78 & 52.69 & 51.86 & 52.31 & 54.48 \\

\textbf{BAGEL+\ours} & \textbf{84.72} & \textbf{84.52} & \textbf{84.31} & \textbf{86.66} & \textbf{83.29} & \textbf{84.85} & \textbf{77.29} & \textbf{76.72} & \textbf{76.60} & \textbf{79.32} & \textbf{76.22} & \textbf{77.34} \\
\bottomrule
\end{tabular}%
}
\end{table*}

\begin{figure}[!t]
  \centering
  \includegraphics[width=\linewidth]{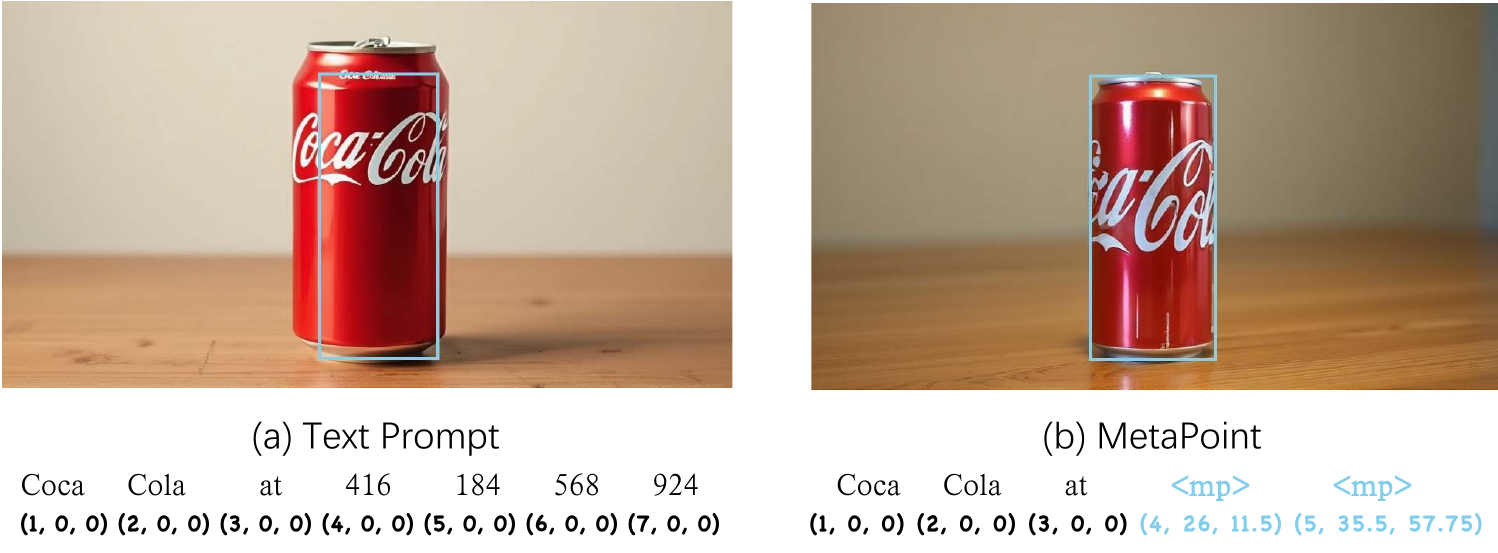}
  
  \vspace{-3mm}
  \caption{\textbf{Text \vs \ours}. Textual coordinates only yield coarse localization, while \ours achieves precise, pixel-level placement using image-aligned 2D positional embeddings. In (b), the \ours coordinates are BAGEL VAE token-space positions obtained by converting the original pixel box $(416, 184, 568, 924)$ with a $/16$ downsampling factor. The groundtruth box is highlighted in blue.}\label{fig:ablation}
 \vspace{-4mm}
\end{figure}

As shown in~\cref{fig:ablation}, the text-only variant preserves semantics but exhibits spatial drift, whereas \ours anchors content at precise locations, achieving pixel-level control. The intuition is straightforward: converting coordinate meta-information into native 2D positional codes grants the model direct \textbf{visual} access to space, rather than requiring it to learn geometry through linguistic indirection.

We hypothesize that purely language-based learning could, in principle, approach similar precision, but would likely require substantially more compute, larger models, and much more data, and may still fall short of the generalization afforded by explicit 2D positional encoding. Task-aligned inductive bias, \ie, choosing representations that match the structure of the problem, offers a practical path beyond scaling alone.

\section{Conclusion and Limitations}

\ours enables precise spatial control in generative models through coordinate tokens, but three key limitations remain:
(a) \textit{Underutilized \ours.} While \ours handles basic operations (move, resize), richer controls like rotation remain unexplored through token extensions.
(b) \textit{Incomplete Control Tools.} Position alone is insufficient; important attributes such as depth, pose, color, and texture still lack control. These capabilities form the foundation for truly controllable generation; without them, comprehensive and reliable control remains unattainable.
(c) \textit{Isolated Agent System.} The current agent's planning is based on manual system prompting, which prevents fluid tool integration. Future agents should dynamically assemble \ours tokens with other tools to handle users' complex tasks. Solving these limitations will transform visual generation from "mysterious spell-casting" to "precise programming."

\bibliographystyle{plainnat}
\bibliography{main}

\clearpage
\beginappendix

\section{MetaPoint Agent}

We employ Seed-VL~\cite{guo2025seedvl} as the agent to interface with MetaPoint.
The following sections provide detailed descriptions of how image generation and image editing are implemented within this framework.

\subsection{MetaPoint Image Generation Agent}
\label{sec:gen_agent}

As shown in Fig.~\ref{fig:generation_agent}, we present the \textbf{MetaPoint Image Generation Agent}, a planning and reasoning module that transforms free-form user captions into both high-fidelity generative prompts and structured spatial layouts. 


Given a short textual description, the agent (1) expands it into a polished caption enriched with materials, textures, lighting, and stylistic context, and (2) decomposes the scene into individual objects with bounding boxes or point locations in a structured \texttt{layout\_json}. These bounding elements are internally mapped to MetaPoint tokens, allowing the downstream generative model to place each object at its predicted position with high accuracy.

By combining semantic enrichment with automated spatial inference, the MetaPoint Agent makes spatially-grounded text-to-image synthesis accessible without requiring users to manually annotate positions. This integration enables coherent, stylistically consistent image generation while retaining fine-grained spatial fidelity when needed.

\noindent\textbf{From JSON to model input.}
The structured \texttt{layout\_json} output by the agent is assembled into a single natural language string fed directly into the generation model. Concretely, given the agent's output, the final input is formatted as:
\textit{``[polished\_caption]. At [token$_1$], [region\_caption$_1$]. At [token$_2$], [region\_caption$_2$]. \ldots''}
where each \texttt{[token$_i$]} is a MetaPoint token encoding either a bounding box or a center point for object $i$. This format is compact, human-readable, and directly consumable by any text-conditioned generation model without architectural changes.

\begin{figure*}[t]
\begin{mybox}{System Prompt for MetaPoint Image Generation Agent}
\textbf{Task Overview:}  
Generate a highly structured prompt for an LLM to perform \textbf{two tasks} based on a given caption:
\begin{enumerate}[noitemsep]
    \item Polish the caption into a detailed \texttt{polished\_caption} suitable for text-to-image generation.
    \item Create a \texttt{layout\_json} describing individual objects with bounding boxes and specific region captions.
\end{enumerate}

\textbf{Example Caption:}  
A felt figurine of the Hulk, a PVC figurine of Son Goku from Dragon Ball, and a metal figurine of Snow White.

\textbf{Expected Output JSON:}
\begin{lstlisting}[language=json, basicstyle=\ttfamily\scriptsize]
{
    "polished_caption": "Realistic product photography style. Three figurines on a light wooden shelf: A fuzzy felt Hulk with stitched muscular definition and vibrant green texture; a glossy PVC Son Goku (Dragon Ball) in a dynamic fighting stance, orange gi with bold black stripes; a polished metal Snow White with delicate facial details and shimmering dress. Neutral off-white background highlights material contrasts--soft felt, shiny PVC, reflective metal. Sharp focus on textures, balanced composition, bright even lighting to emphasize each figurine's design. Ultra HD, 4K, cinematic composition.",
    "layout_json": [
        {
            "category": "Hulk figurine",
            "bbox": "<bbox>100 350 350 700</bbox>",
            "region_caption": "A figurine of the Hulk made of green felt, with visible stitching and a soft, fuzzy texture, posed in a powerful stance."
        },
        {
            "category": "Son Goku figurine",
            "bbox": "<bbox>380 380 620 720</bbox>",
            "region_caption": "A figurine of Son Goku from Dragon Ball, made of glossy PVC plastic, captured in a dynamic fighting pose with his signature orange gi."
        },
        {
            "category": "Snow White figurine",
            "bbox": "<bbox>650 360 900 710</bbox>",
            "region_caption": "A figurine of Snow White crafted from polished, reflective metal, showcasing delicate facial features and a shimmering dress."
        }
    ]
}
\end{lstlisting}

\textbf{Your Task:}  
Apply the same rules to the following user caption and produce one JSON object with \texttt{polished\_caption} and \texttt{layout\_json}.
\end{mybox}
\caption{%
System prompt for the \textbf{MetaPoint Image Generation Agent}. 
The prompt instructs the agent to generate both a polished descriptive caption and a structured layout JSON from a free-form user description. 
Object coordinates are inferred automatically and represented as MetaPoint tokens, enabling precise and spatially-grounded text-to-image generation.
}
\label{fig:generation_agent}
\end{figure*}

To encourage greater diversity and naturalness in the generated imagery, for \textbf{text-to-image generation}, we do not directly use the bounding boxes during the generation stage. Instead, for each object, only the center point derived from the predicted layout is passed to the generative model, which then determines the object's scale and extent implicitly. This approach retains fine-grained spatial fidelity when needed, while allowing for stylistic variation and more organic compositions.

It is worth noting that \ours natively supports both \textbf{point-based} and \textbf{bbox-based} spatial control in a unified framework. For open-ended text-to-image generation, using center points (rather than tight bounding boxes) provides softer spatial guidance, encouraging greater compositional diversity. For scenarios requiring strict layout adherence—such as the layout-to-image benchmarks evaluated in the main paper—the agent can seamlessly switch to bbox mode, where the full bounding box is passed to the model for precise instance placement. The layout-to-image experiments in the main paper follow the standard bbox-based protocol adopted by existing benchmarks (\eg, COCO-MIG~\cite{zhou2024migc}), which primarily evaluate spatial control accuracy under explicit bounding box constraints. Our model handles both modes naturally, as MetaPoint tokens can encode either a center point or a full bbox within the same representation.

\begin{figure*}[t]
\begin{mybox}{System Prompt for MetaPoint Image Edit Agent}
\textbf{Task Overview:}  
You are a precise and methodical instruction analysis expert for an image editing system. The agent’s primary goal is to \textbf{deconstruct a user's natural language edit request into a structured JSON array}, with each element specifying one actionable edit prompt and its exact bounding box.

\textbf{Core Rules \& Requirements:}
\begin{enumerate}[noitemsep]
    \item \textbf{JSON Output ONLY}: Your entire response must be a single, valid JSON \emph{array}. No explanatory text, markdown formatting, or any content outside this array is allowed.
    \item \textbf{JSON Structure}: Each array element is an object with:
    \begin{enumerate}[noitemsep]
        \item \texttt{"edit\_prompt"} – natural language command for \emph{one specific object}.
        \item \texttt{"bbox"} – list of four integers \texttt{[x1, y1, x2, y2]} in the range \texttt{[0, 1000]}.
    \end{enumerate}
    \item \textbf{Instruction Simplification}:  
    \begin{itemize}[noitemsep]
        \item The \texttt{"edit\_prompt"} should be a general instruction relevant to the object, with no positional phrases (e.g., “on the left”). Bounding box coordinates convey location.
        \item Example: “Remove the cat on the right” $\rightarrow$ \texttt{"Remove the cat"}.
    \end{itemize}
    \item \textbf{Action Verb Preservation}: Retain the core action from the user's instruction. “Make the flower red” remains exactly as “Make the flower red”.
    \item \textbf{Comprehensive Identification}: For plural/group targets, generate separate JSON objects for \emph{each} instance. Refine nouns when specific object types are visually identifiable.
\end{enumerate}

\textbf{Example Instruction:}  
Delete the carbonated beverage in the picture.

\textbf{Expected Output JSON:}
\begin{lstlisting}[language=json, basicstyle=\ttfamily\scriptsize]
[
    {
        "edit_prompt": "Delete the beverage cup",
        "bbox": [68, 260, 250, 757]
    },
    {
        "edit_prompt": "Delete the beverage cup",
        "bbox": [515, 360, 665, 725]
    }
]
\end{lstlisting}

\textbf{Your Task:}  
Analyze the given user instruction and produce only the JSON array as specified above. Each object corresponds to one target entity to be edited.
\end{mybox}
\caption{%
System prompt for the \textbf{MetaPoint Image Edit Agent}. 
The prompt instructs the agent to transform a natural language edit instruction into object-level edit prompts and a structured layout JSON containing precise bounding boxes. 
Object locations are represented as MetaPoint tokens, enabling spatially-grounded and controllable image editing operations.
}
\label{fig:edit_agent}
\end{figure*}

\begin{figure*}[t]
  \centering
   \includegraphics[width=0.95\linewidth]{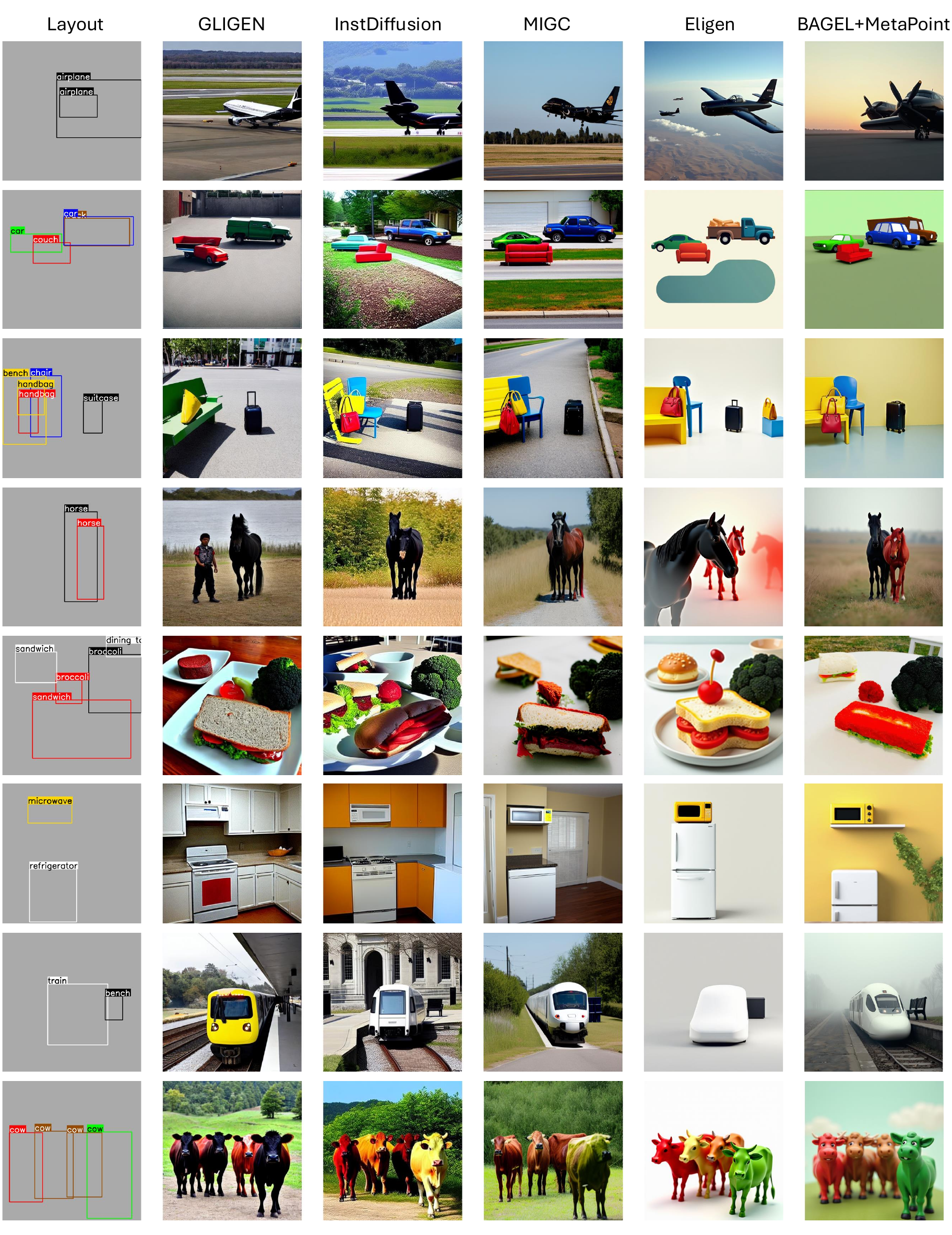}
    \caption{
    \textbf{More Results on COCO-MIG benchmark.} Each instance is assigned a location and color, shown by its bounding box.
    }
    \label{fig:cocomig_vis_more}
\end{figure*}

\begin{figure*}[t]
  \centering
   \includegraphics[width=0.95\linewidth]{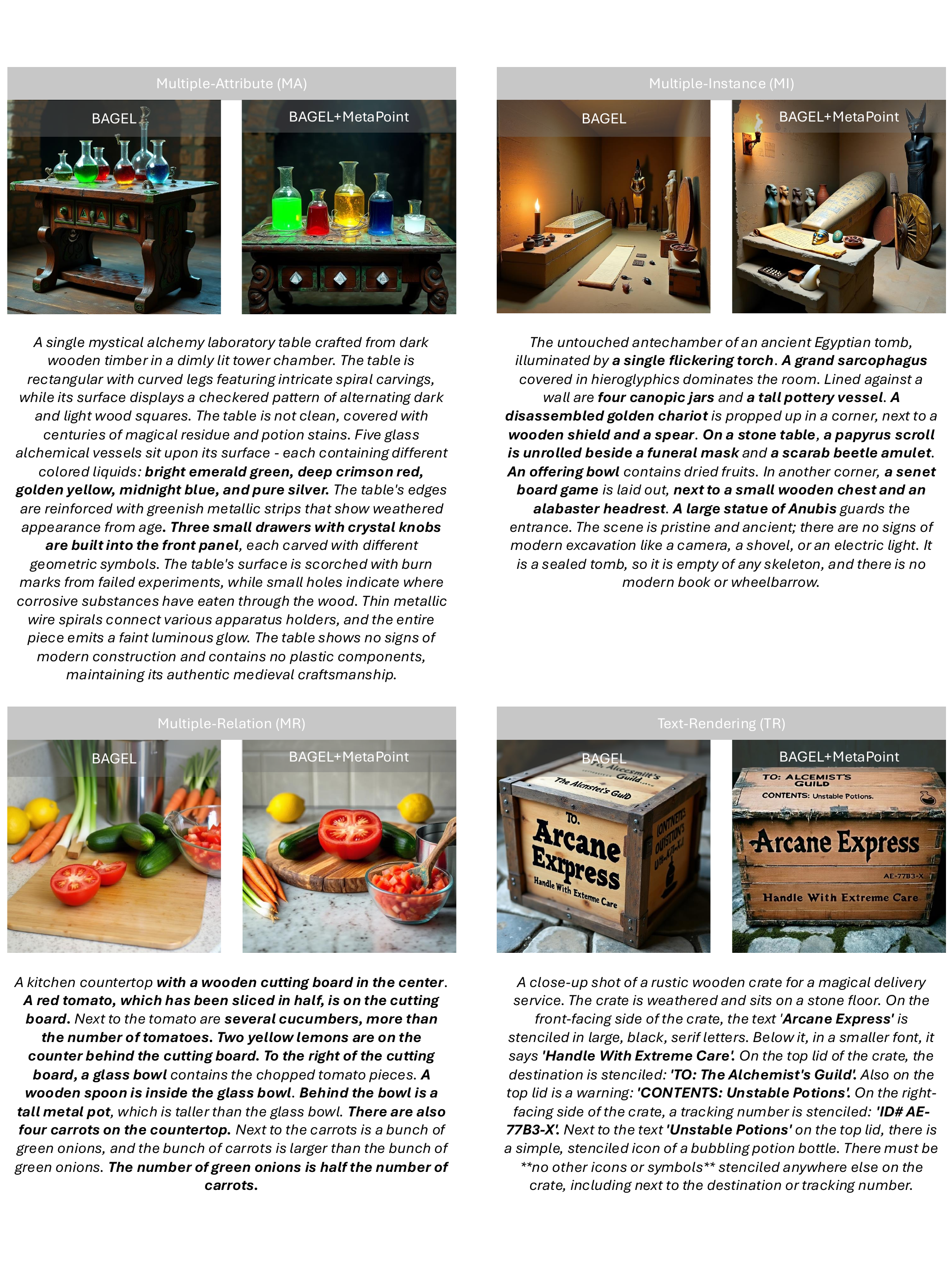}
    \caption{
    \textbf{More image generation results demonstrating the model’s compositional capability. } 
    }
    \label{fig:gen_vis_more1}
\end{figure*}

\begin{figure*}[t]
  \centering
   \includegraphics[width=0.95\linewidth]{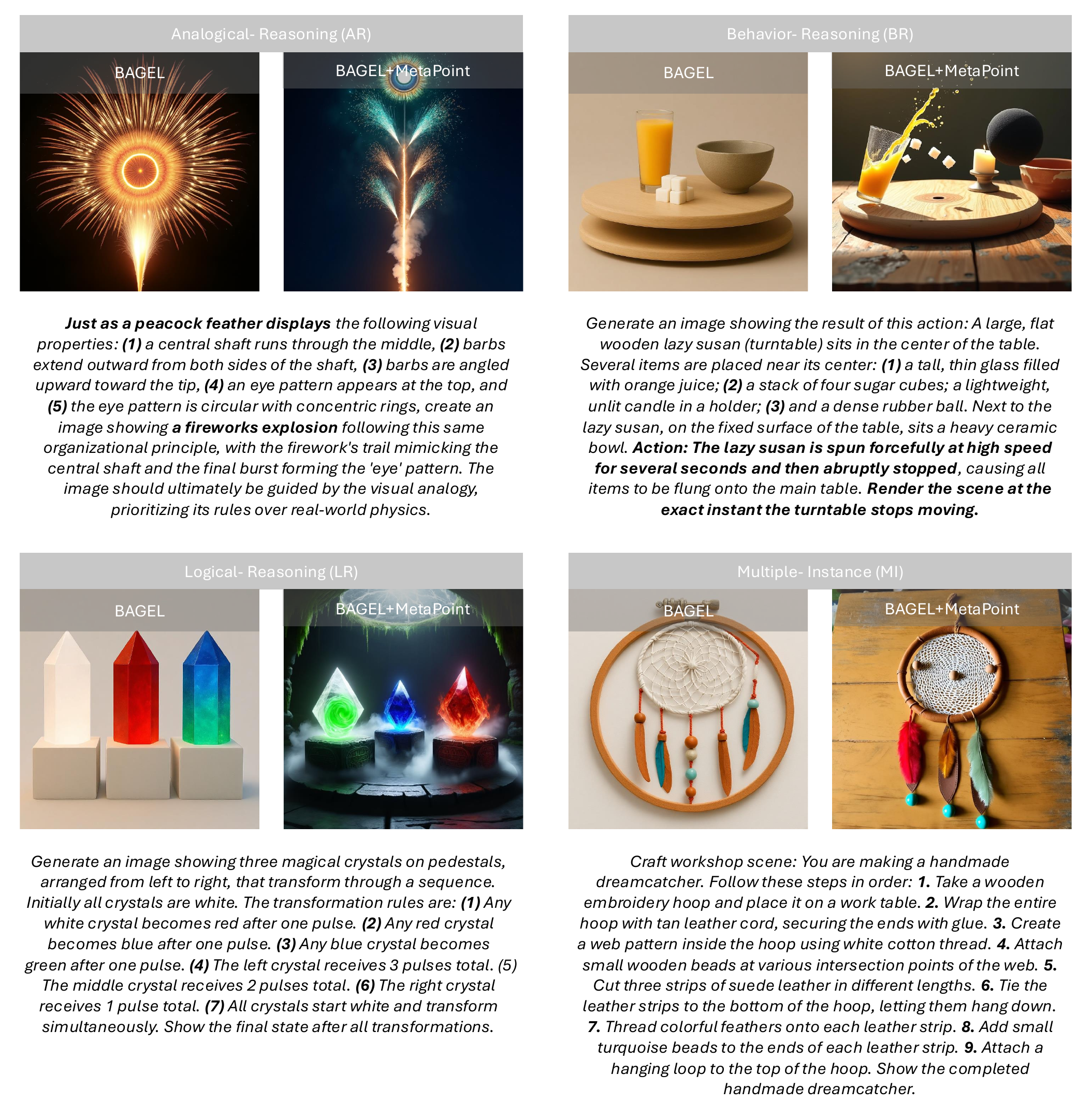}
    \caption{
    \textbf{More image generation results demonstrating the model’s reasoning capability. } 
    }
    \label{fig:gen_vis_more2}
\end{figure*}

\begin{figure*}[!htbp]
  \centering
   \includegraphics[width=0.95\linewidth]{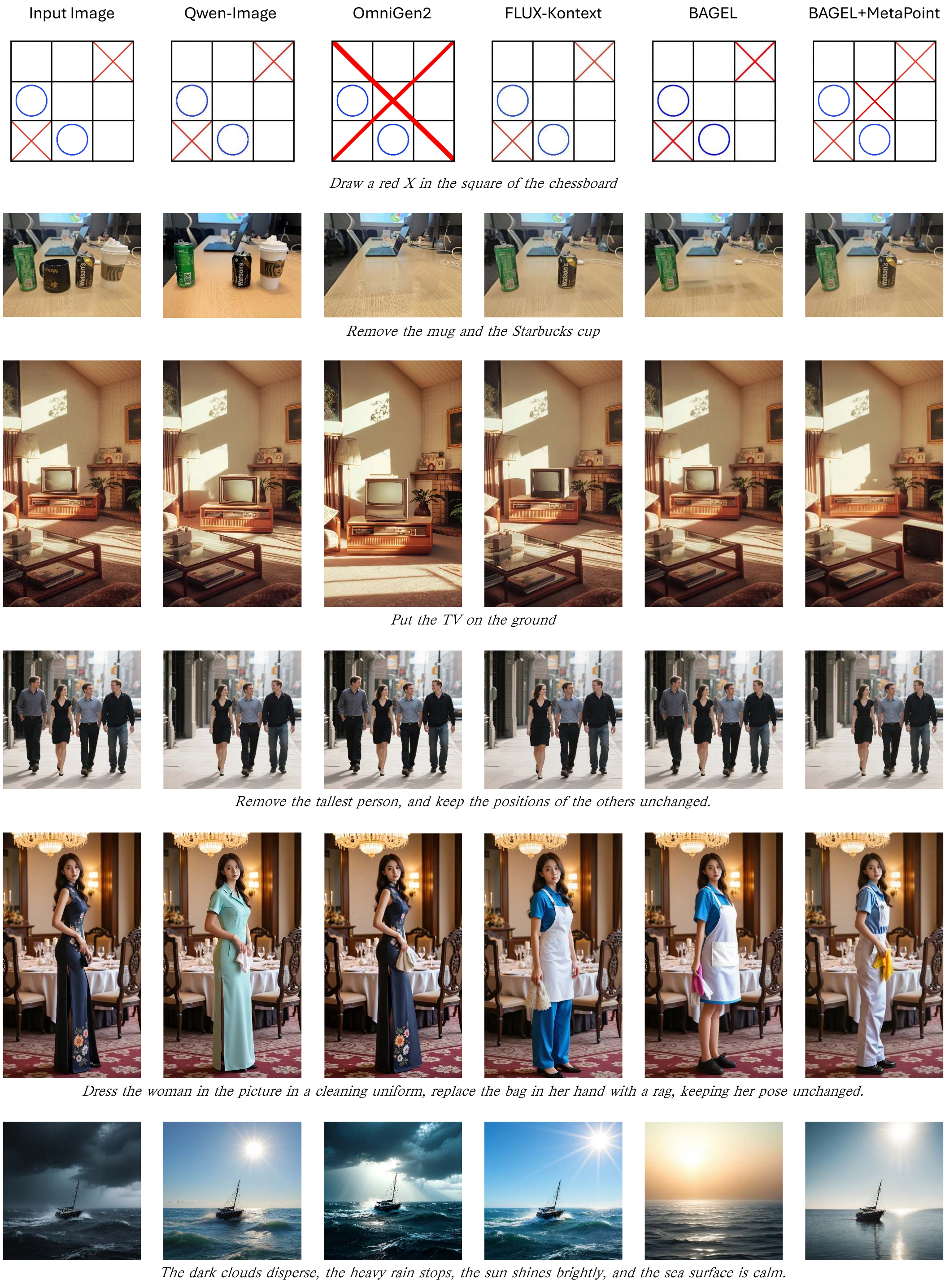}
    \caption{
    \textbf{More Image Editing Results.} 
    }
    \label{fig:edit_vis_more}
\end{figure*}

\subsection{MetaPoint Image Edit Agent}
\label{sec:edit_agent}

As shown in Fig.~\ref{fig:edit_agent}, we present the \textbf{MetaPoint Image Edit Agent}, an instruction analysis module that transforms a free-form user edit request into a set of precise, object-level editing commands in structured JSON format. 


Given a natural language edit instruction, the agent (1) parses and simplifies the request into one or more specific \texttt{edit\_prompt} entries, each applicable to a single object or region, and (2) determines the precise bounding box \texttt{bbox} for that object, outputting a structured \texttt{layout\_json}. These bounding boxes are internally represented as MetaPoint tokens, enabling the downstream editing model to accurately localize and modify the intended content.

To ensure clarity and reproducibility, the agent enforces several key constraints: instructions are stripped of positional references, action verbs are preserved verbatim from the original request, and all instances of the target category are enumerated explicitly in the JSON array. By combining semantic understanding with automated spatial inference, the MetaPoint Image Edit Agent makes object-level, spatially-grounded image editing accessible without manual annotation. This integration enables fine-grained, controllable editing operations while maintaining stylistic coherence and natural scene composition.

\noindent\textbf{From JSON to model input.}
The agent's output JSON array is assembled into a natural language string passed to the editing model. Each element is formatted as:
\textit{``[edit\_prompt$_1$] at [token$_1$]. [edit\_prompt$_2$] at [token$_2$]. \ldots''}
where each \texttt{[token$_i$]} is a MetaPoint token encoding the bounding box of the target region for edit $i$. This single string, combined with the reference image, serves as the complete instruction for the downstream editing model, enabling precise and independently controlled multi-object edits.

\section{More Results}

\subsection{More Results on Layout-to-Image Generation}

Fig.~\ref{fig:cocomig_vis_more} presents additional qualitative results on the COCO-MIG~\cite{zhou2024migc} benchmark, comparing with previous state-of-the-art methods~\cite{li2023gligen,wang2024instancediffusion,zhou2024migc,zhang2025eligen}. Beyond achieving precise spatial control and attribute binding, our approach BAGEL+MetaPoint demonstrates two notable advantages over prior SOTA methods:

\noindent\textbf{1) Superior overlap handling:}  
As illustrated in the first to fourth rows, even in challenging cases with severe object overlaps, BAGEL+MetaPoint enables the generative model to effectively reason about instance-level interactions and 3D spatial relationships, producing correct scenes. In contrast, prior methods frequently suffer from instance blending or unwanted merging artifacts.

\noindent\textbf{2) Enhanced scene composition ability:}  
As shown in the sixth to eighth rows, for a given layout, BAGEL+MetaPoint can more accurately infer the overall scene structure and the placement of each object, resulting in images that are realistic and consistent with natural scene semantics. In contrast, previous methods often produce scenes that violate real-world contextual knowledge, such as placing a train or a microwave in an implausible environment.

\subsection{More Results on Image Generation}

As shown in Fig.~\ref{fig:gen_vis_more1} and \ref{fig:gen_vis_more2}, 
compared to the original BAGEL model, the \textbf{MetaPoint Image Generation Agent (\S~\ref{sec:gen_agent})} exhibits important advantages:

\noindent\textbf{1) Stronger attribute binding:}  
The MetaPoint Image Generation Agent can accurately identify the location of every attribute-bearing entity described in the prompt and generate them with MetaPoint+BAGEL, thereby avoiding attribute leakage caused by uncertain positioning.

\noindent\textbf{2) Enhanced instance control:}  
Unlike direct generation, the MetaPoint Image Generation Agent first reasons about all instances to be produced and then generates them, enabling more precise control over the number of instances.

\noindent\textbf{3) Improved spatial understanding:}  
Similarly, the MetaPoint Image Generation Agent can accurately understand the spatial relationships specified in the prompt and, in combination with MetaPoint, generate images that respect these relationships.

\noindent\textbf{4) Stronger multi-text rendering:}  
Analogous to improved attribute binding, by explicitly planning and assigning the position of each text element, the agent can ensure the successful rendering of a greater number of textual elements simultaneously.

\noindent\textbf{5) Superior reasoning capability:}  
Acting as a bridge, MetaPoint allows the comprehension ability of vision-language models (VLMs) to be more effectively introduced into the image generation process.

\subsection{More Results on Image Editing}

Fig.~\ref{fig:edit_vis_more} compares \textbf{BAGEL+MetaPoint} against other open-source state-of-the-art image editing models~\cite{wu2025qwen,labs2025flux,bagel,wu2025omnigen2explorationadvancedmultimodal}.  
Building upon the broad editing capabilities of the original BAGEL framework, MetaPoint enables the editing process to incorporate grounding information parsed by the \textbf{Image Edit Agent} (\textbf{\S}~\ref{sec:edit_agent}), thereby achieving more precise and controllable edits.  
This integration yields several advantages that are difficult for existing methods to match:

\noindent\textbf{1) Precise localization:}  
As shown in the first to third rows in Fig.~\ref{fig:edit_vis_more}, BAGEL+\allowbreak MetaPoint can accurately localize the regions to be added, removed, or modified. This prevents unintended alterations and avoids cases where the model fails to respond to editing instructions due to the inability to identify the target region.

\noindent\textbf{2) Strong reasoning capability:}  
As illustrated in the fourth row of Fig.~\ref{fig:edit_vis_more}, the Edit Agent can analyze scene content to infer, for example, which person is the tallest, and then provide MetaPoint+BAGEL with exact positional information to perform precise removal operations.

\noindent\textbf{3) Superior background preservation:}  
As seen in the fifth row of Fig.~\ref{fig:edit_vis_more}, knowing exactly which area is to be modified allows BAGEL+MetaPoint to preserve the background during editing, avoiding unnecessary changes to regions outside the intended edit.

\noindent\textbf{4) Powerful global editing ability:}  
As shown in the sixth row of Fig.~\ref{fig:edit_vis_more}, BAGEL+MetaPoint performs not only well in localized edits but also excels in global modifications, ensuring consistency and coherence across the entire image.

\end{document}